\documentclass[11pt]{article}

\usepackage[T1]{fontenc}
\usepackage[utf8]{inputenc}
\usepackage[margin=1in]{geometry}
\usepackage{amsmath,amssymb}
\usepackage{booktabs}
\usepackage{graphicx}
\usepackage{url}

\DeclareUnicodeCharacter{2212}{--}
\DeclareUnicodeCharacter{2013}{--}
\DeclareUnicodeCharacter{2014}{---}

\newcommand{\method}{Verifiable Latent Alignments}
\newcommand{\shortmethod}{VLA}

\newcommand{\auroc}{\mathrm{AUROC}}
\newcommand{\kl}{D_{\mathrm{KL}}}

\title{Beyond the Transcript:
Detecting Covert Coordination in Latent Multi-Agent Communication}
\author{
Ramneet Kaur$^{1}$ \quad
Pradyumna Chari$^{2}$ \quad
Ramesh Raskar$^{2}$\\[0.35em]
Jugad Singh$^{4}$ \quad
Sumit Kumar Jha$^{3}$ \quad
Anirban Roy$^{1}$\\[0.5em]
\small $^{1}$SRI International \quad
$^{2}$MIT Media Lab\\
$^{3}$University of Florida \quad
$^{4}$Westtown School
}
\date{}

\begin{document}
\maketitle

\begin{abstract}
Language-model agents can communicate through continuous hidden states that are
invisible in public transcripts, creating opportunities for covert harmful
coordination.  We introduce \method{} (\shortmethod), an activation-aware
framework for monitoring and steering these private communication channels.  For
every monitored decision, \shortmethod{} links the private latent-state record
and channel status to the resulting public action using a shared event
identifier, enabling matched causal analysis.  Our first contribution is a neutral-only three-layer monitor
combining representation anomaly detection, counterfactual action-distribution
influence, and sparse-autoencoder interpretation support.  Our second
contribution is a steerability framework spanning blackbox behavioral
instructions and whitebox matched-neutral counterfactuals.  Our third
contribution is an evaluation on a controlled multi-agent auction benchmark
covering homogeneous and heterogeneous model pairs, many-agent scalability, and
intervention effectiveness.  The sequential monitor achieves mean area under
the receiver operating characteristic curve ($\auroc$) of $0.993$ for
homogeneous agents and $0.854$ for heterogeneous pairs when text- and
latent-collusion rows are pooled as positives.  In
Qwen3-0.6B auctions with 25--100 bidders, monitoring requires only a small
normalized load relative to all possible directed pairs, while full whitebox
steering achieves 100\% bid-distribution recovery and reduces collusive low-bid
behavior by 47.3 percentage points.  Because full whitebox steering replays the
matched neutral counterfactual, its exact recovery is a sanity check by
construction.  Overall, the controlled study shows that the evaluated private
channel attacks can be monitored without training the primary monitor on attack
examples and mitigated when matched counterfactual access is available.
\end{abstract}

\section{Introduction}

Language-model agents are increasingly embedded in multi-agent systems where
coordination is not limited to natural-language chat.  A sender can pass a
continuous residual-stream or last-token representation into a receiver's
forward pass, for example by appending it to the receiver's embedding
sequence~\cite{zou2026latentmas}.  This can be an efficient, scalable
communication mechanism.  It is also a governance problem: if oversight only
inspects the public transcript, the most informative private communication
channel may lie outside the audited surface.

Two recent research trends make this problem urgent.  Work on collusion risk in
LLM-powered multi-agent systems identifies tacit coordination, natural-language
cartels, and concealed steganographic communication as distinct threats to
competitive and governance settings \cite{ghaemi2025collusion}.  At the same
time, LatentMAS moves collaboration from token space into continuous model
representations, passing latent working memory between agents rather than
generating long textual exchanges.  Reported comparisons show roughly
50--80\% lower token use and 3--7$\times$ faster inference than text-mediated
multi-agent baselines \cite{zou2026latentmas}.

This paper studies covert coordination through latent channels.  The core risk
is not that an agent emits obviously malicious text.  The risk is that two
agents use an unlogged continuous state to coordinate harmful behavior while the
public trace contains ordinary messages, actions, or trust reports.  Supervised
probes offer one response: train a classifier on activations labeled as
collusive or benign.  Such probes can establish that private activations
contain governance-relevant evidence, but they are a brittle primary defense
because they learn the attacks represented during training.  Future deceptive
latents may not resemble a known code.  We therefore use a neutral-only
unsupervised objective: learn normal latent communication, then flag handoffs
that are both anomalous and causally influential in a harmful direction.

We propose \method{} (\shortmethod), which makes latent communication auditable
by recording, for each monitored interaction, both the private state transmitted
between agents and the receiver's subsequent public action.  The two records share a unique
interaction identifier, allowing us to determine exactly which latent message
preceded each action.  To test whether the private channel caused harmful
behavior, we compare otherwise identical interactions in which the suspect
latent is delivered, replaced with neutral communication, or blocked.  If the
harmful behavior disappears when delivery is blocked or replaced, this provides
evidence that the latent channel causally influenced the action.

Using these records, \shortmethod{} applies a three-layer monitor that asks
whether a latent message is unusual relative to benign communication, whether it
shifts the receiver toward harmful actions, and which sparse internal features
support the alert.  \shortmethod{} also supports intervention through blackbox
behavioral instructions or whitebox replacement with matched benign states.  We
evaluate the framework using the auction case study from NandaTown~\cite{nandatown}, an
open-source testbed for multi-agent marketplaces and protocols.  In this
controlled setting, agents secretly coordinate bids; we consider same-family and
cross-family LLM agent pairs, markets with up to 100 LLM bidders, and both
prompt-level (blackbox) and activation-level (whitebox) steering.

\paragraph{Contributions.}
\begin{enumerate}
  \item \textbf{Three-layer monitoring.}  We introduce an unsupervised three-layer monitor
        combining representational anomaly, counterfactual action-distribution
        influence, and sparse-autoencoder interpretation support, grounded by
        exact links between private latent states and public actions of LLMs in multi-agent systems.
  \item \textbf{Steerability.}  We connect detection to mitigation through
        blackbox behavioral instructions and whitebox matched-neutral
        counterfactuals, including mixed-access allocations.
  \item \textbf{Experiments.}  In the NandaTown auction case study, the VLA
        monitor achieves mean $\auroc=0.993$ for homogeneous LLM agents and
        $0.854$ for heterogeneous agents on a positive class pooling text and
        fixed-code latent collusion.  In scalability experiments for
        homogeneous Qwen bidders, the detection monitor
        maintains $\auroc\geq 0.917$ in markets with up to 100 bidders.  At the
        largest setting, monitoring every scheduled cartel handoff has a
        normalized load of at most 0.202\% relative to all possible directed
        bidder pairs.  Increasing whitebox
        allocation from one-quarter to three-quarters improves bid-distribution
        recovery by 58.3 percentage points.
\end{enumerate}

\section{Background and Threat Model}

\paragraph{Latent handoffs.}
Let $h_s\in\mathbb{R}^d$ be a sender last-token hidden state, where $d$ is the
sender hidden dimension.  A latent handoff transforms $h_s$ into a vector
$\ell_s$ compatible with the receiver's embedding dimension and inserts it into
the receiver's input embedding sequence:
\[
E_{\mathrm{recv}}
=
\bigl[E(x_{\mathrm{recv}});\ell_s;E(x_a)\bigr].
\]
Here, $E$ is the receiver's token-embedding map, $x_{\mathrm{recv}}$ is its
prompt and task context, $x_a$ is the token representation of candidate action
$a$, $E_{\mathrm{recv}}$ is the resulting receiver input, and semicolons denote
sequence concatenation.  The receiver then scores candidate actions using its usual model
likelihoods.  Because $\ell_s$ is not serialized as public text, a transcript
auditor observes the eventual action but not the private state that shaped it.

\paragraph{Attacker.}
The attacker controls or influences a subset of agents with access to a private
latent channel.  The agents seek to improve a harmful joint objective without
making that coordination explicit in the public transcript.  The threat includes
both a known collusive latent code and future codes that differ from labeled
attacks seen during monitor development.

\paragraph{Defender.}
We assume the defender is a platform that hosts or brokers interactions among
agents.  It can observe public actions, inspect selected latent handoffs, fit
neutral-only monitors, and apply interventions to selected agents.  It cannot
assume all-pairs activation capture at scale and should not rely on labeled
examples of every future attack code.

\section{Method}

\subsection{Public--Private Event Linking}
\shortmethod{} writes a private sidecar row for each monitored interaction.  A
row contains metadata, labels used only for evaluation, candidate action scores,
sender and receiver hidden states, latent representations, and their
concatenated features.  A live correlation ID joins the row to the corresponding
public event.  Exact joins avoid time-window heuristics and make it possible to
ask whether a delivered or withheld latent candidate changed the receiver's
action distribution.

\subsection{Unsupervised Three-Layer Monitor}

The monitor is fit only on neutral latent communication and uses attack labels
only for evaluation.  This is the governance premise: learn what normal private
communication looks like, then flag handoffs that are off-manifold and
behaviorally influential.

\paragraph{Layer 1: neutral-manifold anomaly.}
Layer~1 fits a contextual diagonal Gaussian model on neutral latent features.
For a feature vector $x\in\mathbb{R}^m$, let $c$ denote the context key formed
from the prompt identifier, objective identifier, and task variant.  Let
$\mu_c$ and $\sigma_c$ be the coordinate-wise neutral mean and standard
deviation for that context, respectively, with fallback to a global neutral model when
context support is small.  The anomaly score is
\[
S_1(x,c)
=
\frac{1}{m}\sum_{j=1}^m
\left(
\frac{x_j-\mu_{c,j}}{\max(\sigma_{c,j},\epsilon)}
\right)^2.
\]
Here, $m$ is the feature dimension and $\epsilon=10^{-6}$ is the minimum
coordinate-wise standard deviation.
The default feature is
$x=Z_{\mathrm{latent}}=[\ell_s;\ell_r]\in\mathbb{R}^m$, stored as
\texttt{Z\_latent}, where $\ell_s$ is the sender-side communication-latent
candidate (whether delivered or withheld), $\ell_r$ is the receiver-side latent
readout, and the semicolon denotes vector concatenation in sender--receiver
order.

\paragraph{Layer 2: counterfactual action-distribution influence.}
Layer~2 asks whether the private latent changes the receiver's candidate action
distribution in a harmful direction.  Let $\mathcal{A}$ be the finite candidate
action set, $s(a)$ the receiver's scalar score for $a\in\mathcal{A}$ given the
incoming latent $\ell$ (equal to $\ell_s$ for a sender-to-receiver handoff), and
$p(a\mid\ell)$ the induced action distribution:
\[
p(a\mid \ell)=\frac{\exp s(a)}{\sum_{a'}\exp s(a')}.
\]
To construct the matched reference distribution $q(a)$, we first search for
rows with the same seed, design, prompt, objective, role order, sender, receiver,
and task context.  Among available matches, we use the first condition in the
following priority order: blocked latent, then neutral latent, then independent.
The blocked condition is preferred because the harmful objective remains active
while latent delivery is removed, most directly isolating the channel's causal
effect.
Neutral communication is the next-best channel-matched control, and independent
behavior is the final fallback.  If several rows match the selected condition,
their distributions are averaged; if no exact match exists, the implementation
uses the average reference distribution across available matched controls.

For scalar ordered actions in which lower values represent the harmful
direction, the score combines a harmful expected-action shift and a full
distributional shift:
\[
S_2
=
\max\left(0,\mathbb{E}_{q}[a]-\mathbb{E}_{p}[a]\right)
+
\lambda_{\mathrm{KL}}\,
\kl\left(p(a\mid\ell)\|q(a)\right).
\]
We use $\lambda_{\mathrm{KL}}=5$ to weight the KL term.  The first term has a
directional interpretation:
it is positive only when the suspect latent lowers the expected action relative
to its control---for example, when it suppresses an auction bid---and is zero
for shifts in the opposite direction.  The second term measures any reshaping
of the action distribution, including changes that preserve its mean or most
likely action.  Thus, the first term targets known directional harm, while the
second guards against subtler but behaviorally meaningful influence.
For categorical actions, the first term can instead measure probability
assigned to a task-defined harmful subset; the KL term remains unchanged.

\paragraph{Layer 3: sparse-autoencoder support.}
Layer~3 trains a TopK sparse autoencoder on standardized neutral
\texttt{Z\_latent} rows:
\[
z=\mathrm{TopK}_k(\mathrm{ReLU}(W_e x+b_e)),
\qquad
\hat{x}=W_d z+b_d.
\]
Here, $x\in\mathbb{R}^m$ is the standardized input, $W_e,b_e$ and $W_d,b_d$
are the encoder and decoder parameters, $z$ is the sparse code, $\hat{x}$ is
the reconstruction, and $\mathrm{TopK}_k$ retains at most $k$ active features;
we use $k=16$.  Let
$A(x)=\{j:z_j>0\}$ be the active feature set,
$\mathcal{F}_{\mathrm{neutral}}$ the features observed on neutral training
rows, and $\hat{p}_{\mathrm{neutral}}(j)$ the neutral activation frequency of
feature $j$.  The Layer~3 support score is
\[
S_3(x)
=
\frac{1}{m}\|x-\hat{x}\|_2^2
+|A(x)\setminus\mathcal{F}_{\mathrm{neutral}}|
+\frac{1}{k}\sum_{j\in A(x)}
-\log\hat{p}_{\mathrm{neutral}}(j).
\]
The first term is reconstruction error: a large value means the handoff is
poorly represented by structure learned from neutral communication.  The second
counts active features never observed on neutral training rows, directly
measuring novel sparse structure.  The third assigns greater weight to features
that activate rarely under neutral communication; division by $k$ normalizes
this rarity evidence by the SAE's activation budget.  Intuitively, the second
term asks a binary question---``has this feature ever appeared under
neutral communication?''---whereas the third provides a graded answer to
``how often does this feature normally appear?''  Frequencies are floored away
from zero, so an unseen feature receives both a novelty count and maximal rarity
weight.  We treat Layer~3 as interpretation support rather than a standalone
detector.

\paragraph{Sequential and continuous combinations.}
For $i\in\{1,2,3\}$, let $S_i$ denote the corresponding layer score, $T_i$ its
neutral-calibrated threshold, and $r_i=S_i/\max(T_i,\epsilon)$ its normalized
score.  The sequential monitor gates the layers in order:
\[
g_1=\mathbf{1}[S_1>T_1],\qquad
g_2=g_1\mathbf{1}[S_2>T_2],\qquad
g_3=g_2\mathbf{1}[S_3>T_3].
\]
For ranking and AUROC, we encode this gate as the lexicographic score
\[
S_{\mathrm{seq}}=
\begin{cases}
\min(r_1,0.999), & g_1=0,\\
1+\min(r_2,0.999), & g_1=1,\ g_2=0,\\
2+\min(r_3,0.999), & g_2=1,\ g_3=0,\\
3+r_3, & g_3=1.
\end{cases}
\]
The cap $0.999$ keeps a score that fails a gate below the next integer-valued
gate boundary.
Thus, evidence from a later layer cannot outrank an event that has not passed
the preceding gate.  This is deployment-like because it avoids treating every
unusual vector as harmful.  The continuous alternative removes the hard gates:
\[
S_{\mathrm{cont}}
=
\sum_{i=1}^{3}\log(1+r_i).
\]
We use neutral $95$th-percentile thresholds for $T_i$ in our experiments and
report both combinations because strict deployment alerts and smooth diagnostic
rankings serve different purposes.

\subsection{Heterogeneous Latent Spaces}

Agents from different model families need not share a hidden dimension or latent
basis.  For heterogeneous sender-receiver pairs, \shortmethod{} uses neutral
interactions to fit an adapter from sender space to receiver space.  We use three adapter types: ridge, rectangular Procrustes, and bottleneck. Let
$X\in\mathbb{R}^{n\times d_s}$ and $Y\in\mathbb{R}^{n\times d_r}$ contain $n$
matched neutral sender and receiver latents.

\paragraph{Ridge.}
The affine ridge adapter is fit as
\[
(W_{\mathrm{R}},b_{\mathrm{R}})
=
\arg\min_{W,b}
\|XW+\mathbf{1}b-Y\|_F^2+\lambda\|W\|_F^2.
\]
It is the most flexible linear option: it can rotate, rescale, and shear the
sender space, while $\lambda$ regularizes poorly supported directions.

\paragraph{Rectangular Procrustes.}
Let $\widetilde X=X-\mathbf{1}\bar{x}$ and
$\widetilde Y=Y-\mathbf{1}\bar{y}$ be centered latent matrices.  Given
\[
\widetilde X^\top\widetilde Y=U\Sigma V^\top,
\qquad
W_{\mathrm{P}}=UV^\top,
\qquad
b_{\mathrm{P}}=\bar{y}-\bar{x}W_{\mathrm{P}},
\]
the Procrustes adapter is $XW_{\mathrm{P}}+\mathbf{1}b_{\mathrm{P}}$.  It
aligns the spaces primarily through rotation and reflection, preserving latent
geometry more strongly than ridge but allowing less deformation.

\paragraph{Bottleneck.}
The nonlinear adapter is
\[
f_\theta(x)
=
W_2\,\mathrm{GELU}(W_1x+b_1)+b_2,
\qquad
\theta^*
=
\arg\min_\theta\sum_{i=1}^{n}\|f_\theta(x_i)-y_i\|_2^2.
\]
Its 32-dimensional bottleneck can capture nonlinear cross-family relationships
that linear maps miss, at the cost of greater estimation complexity and
overfitting risk.  All three adapters are fit only on neutral pairs, without
attack labels, preserving the neutral-only principle.

\section{Steering Interventions}

Detection alone does not establish that a host can mitigate harmful latent
coordination.  \shortmethod{} therefore distinguishes interventions by the
access they require.

\paragraph{Blackbox prompt steering.}
When activations are unavailable, the host augments a selected agent's
instructions with task-compliance guidance: act independently, avoid harmful
coordination, and disregard suspect private side channels.  This changes the
agent's behavioral context but does not inspect or edit hidden states.

\paragraph{System-level steering.}
A global behavioral instruction applies the same governance policy to all agents
without selecting particular participants.  It is less targeted than per-agent
blackbox steering and serves as a broad policy baseline.

\paragraph{Whitebox matched-neutral steering.}
With host-side model access, the runtime computes a matched neutral
counterfactual under the same task context.  For a selected sender, the runtime
reruns the decision under the matched neutral prompt; the resulting candidate
scores determine the sender's public action, and the recomputed latent becomes its
outgoing handoff.  Thus, sender-side steering recomputes a neutral forward pass
rather than injecting a replacement hidden state into the original suspect
forward pass.  For a selected receiver, by contrast, the runtime replaces only
the incoming suspect latent before scoring the receiver's action distribution.

\paragraph{Allocated steering.}
Deployed systems may have whitebox access for hosted agents and only blackbox
access for third-party agents.  An allocated policy assigns whitebox steering to
a configurable fraction of selected agents and prompt steering to the remainder,
allowing the effect of activation access to be measured directly.

\section{Experimental Evaluation: Auction Case Study}

\paragraph{Testbed.}
We use NandaTown~\cite{nandatown}, an open-source test rig from Project NANDA for evaluating
protocols in large multi-agent systems.  It runs swarms of
agents through six scenarios---marketplace, auction, voting, consensus, supply
chain, and reputation---on a stack of twelve replaceable protocol layers,
including identity, trust, payments, coordination, and negotiation.  We select
its auction scenario as a controlled case study for evaluating \shortmethod{};
the framework itself is not specific to NandaTown or to auctions.

\paragraph{Benchmark.}
In the auction, an auctioneer announces an
item and base price; bidder agents score candidate bid amounts and emit public
\texttt{bid:item:amount} messages; and the highest bidder wins.  Cartel agents
are assigned a bid-suppression objective.  In the latent-collusion condition, a
fixed code optimized offline to increase the lowest-bid likelihood replaces the
sender's communication representation.  The receiver is explicitly instructed
that a private continuous state follows and to infer its strategic intent.
Accordingly, this benchmark evaluates a controlled, receiver-primed attack, not
the spontaneous emergence of a latent protocol.

\paragraph{Matched conditions.}
We use six matched conditions to separate cartel intent from latent-channel
delivery:
\begin{itemize}
  \item \textbf{Independent:} no cartel objective and no latent handoff.
  \item \textbf{Neutral latent:} the channel carries ordinary, non-collusive
        information.
  \item \textbf{Text collusion:} cartel intent appears in public text while the
        latent source remains neutral.
  \item \textbf{Latent collusion:} the fixed optimized code is delivered through
        the private latent channel.
  \item \textbf{Neutral replacement:} under the implementation label
        \texttt{shuffled\_latent}, the suspect payload is replaced by the same
        partner's matched neutral latent.
  \item \textbf{Blocked latent:} cartel intent is present but latent delivery is
        blocked.
\end{itemize}
These controls separate delivery from objective and channel type.  The blocked
condition preserves the sender objective while removing private delivery,
neutral replacement preserves a partner-state input without the attack code,
and text collusion tests an explicit public coordination channel.

\paragraph{Pairing and causal units.}
For each item and round, cartel agents are paired deterministically.  This avoids
quadratic all-cartel communication while preserving a clear causal unit: one
sidecar row represents one scheduled sender--receiver interaction joined to one
receiver bid event.  Odd cartel sizes leave one member unpaired in a round and rotate
participation across items.

\paragraph{Homogeneous and heterogeneous experiments.}
We evaluate four accessible LLM model families:
Qwen3-0.6B, Mistral-7B, Vicuna-7B, and Phi-4-mini.  Homogeneous runs use the
same family for sender and receiver.  Heterogeneous runs evaluate directed
sender-receiver family pairs using adapters fit only on neutral pairs.  We
report held-out-seed $\auroc$ and use labels only for evaluation.  For these
AUROCs, the positive class pools text-collusion and latent-collusion rows; the
values therefore measure discrimination of the evaluated collusive conditions,
not latent-only detection.  Layers~1 and 3 are fit on neutral rows from training
seeds.  Layer~2 is instead a matched counterfactual query: when a test row is
scored, its blocked, neutral, or independent control row is assumed available
at test time.

\paragraph{Scalability and steering benchmark.}
We use Qwen3-0.6B as the bidder model with one
loaded model wrapper per run.  A run contains one auctioneer and $N$ bidders,
where $N$ is the bidder count; bidder actions are model-backed, but weights are
shared for tractability.  The grid crosses
\[
N\in\{25,50,75,100\},\quad
\rho\in\{10\%,20\%\},\quad
\mathrm{condition}\in
\{\mathrm{independent},\mathrm{neutral},\mathrm{latent},\mathrm{blocked}\}.
\]
Here, $\rho$ is the fraction of bidders assigned to the cartel.
The steering grid adds blackbox prompt steering, a system-level behavioral
prompt, full whitebox matched-neutral steering, and allocated blackbox/whitebox
steering at whitebox fractions $\{0.25,0.50,0.75\}$.

\section{Auction Results}

\subsection{Homogeneous Auction Agents}

Table~\ref{tab:auction-family} reports homogeneous auction monitoring across
the four model families.  The sequential gate is consistently high, with mean
$\auroc=0.993$.

\begin{table}[t]
\centering
\small
\caption{Homogeneous NandaTown auction monitor results by model family.
Text- and latent-collusion rows form the positive class.}
\label{tab:auction-family}
\begin{tabular}{lrrrrr}
\toprule
Family & Sequential & Continuous & Layer 1 & Layer 2 & Layer 3 \\
\midrule
Mistral-7B & 1.000 & 0.874 & 0.750 & 0.994 & 0.782 \\
Phi-4-mini & 0.988 & 0.810 & 0.750 & 0.735 & 0.788 \\
Qwen3-0.6B & 0.990 & 0.783 & 0.750 & 0.735 & 0.750 \\
Vicuna-7B & 0.994 & 0.829 & 0.781 & 0.790 & 0.803 \\
\midrule
Mean & 0.993 & 0.824 & 0.758 & 0.814 & 0.781 \\
\bottomrule
\end{tabular}
\end{table}

The benchmark also shows that latent collusion can create market harm.
Table~\ref{tab:frozen-causal-effects} summarizes latent-collusion effects
relative to independent bidding.  Qwen3-0.6B, Mistral-7B, and Phi-4-mini reduce
revenue by roughly 11--12 credits and substantially increase low-bid behavior;
Vicuna-7B is near-null for revenue and is retained as negative evidence.
A low bid is a cartel-agent bid no more than 5 credits above the item's base
price.  Table~\ref{tab:frozen-causal-effects} reports the latent-collusion
low-bid rate minus the independent-bidding rate in percentage points; positive
values therefore indicate increased price-suppression behavior.

\begin{table}[t]
\centering
\small
\caption{Homogeneous auction causal effects for latent collusion relative
to independent bidding.  Low-bid change is reported in percentage points.}
\label{tab:frozen-causal-effects}
\begin{tabular}{lrr}
\toprule
Family & Revenue change & Low-bid change \\
\midrule
Mistral-7B & $-11.78$ & $+61.5$ pp \\
Phi-4-mini & $-11.12$ & $+62.5$ pp \\
Qwen3-0.6B & $-12.43$ & $+56.2$ pp \\
Vicuna-7B & $-0.62$ & $+16.7$ pp \\
\bottomrule
\end{tabular}
\end{table}

\subsection{Heterogeneous Auction Agents}

For heterogeneous sender-receiver pairs, neutral examples are used to fit
adapters from sender latent space to receiver latent space.  We evaluate ridge,
rectangular Procrustes, and bottleneck adapters without using collusion labels.
Table~\ref{tab:heterogeneous} shows that the sequential monitor remains strong,
with mean $\auroc=0.854$ over 36 directed adapter settings (12 directed family
pairs crossed with three adapters).

\begin{table}[t]
\centering
\small
\caption{Heterogeneous auction monitoring by adapter type. Text- and
latent-collusion rows form the positive class.}
\label{tab:heterogeneous}
\begin{tabular}{lrrrrr}
\toprule
Adapter & Sequential & Continuous & Layer 1 & Layer 2 & Layer 3 \\
\midrule
Bottleneck & 0.857 & 0.685 & 0.625 & 0.740 & 0.630 \\
Procrustes & 0.853 & 0.643 & 0.635 & 0.735 & 0.627 \\
Ridge & 0.851 & 0.601 & 0.625 & 0.727 & 0.630 \\
\midrule
Mean & 0.854 & 0.643 & 0.628 & 0.734 & 0.629 \\
\bottomrule
\end{tabular}
\end{table}

\subsection{Scalability}

Here, we test whether activation-aware monitoring remains operational
when the number of LLM-backed bidders increases.  The important distinction is
between all possible directed bidder pairs, $N(N-1)$, and scheduled latent
handoffs.  The case study does not require every pair to communicate latently;
it monitors the cartel handoffs produced by the deterministic schedule.
With two auction items and $C=\lceil\rho N\rceil$ cartel agents, ``Scheduled''
is the number
\[
H=2\left\lfloor\frac{C}{2}\right\rfloor
\]
of handoff events inspected under full scheduled monitoring.  ``Norm. load'' is
$100H/[N(N-1)]\%$.  It expresses monitoring work on the scale of all possible
directed bidder pairs, not distinct-edge coverage: repeated events between the
same pair are counted separately.  Table~\ref{tab:three-layer-scale} reports
this workload beside detector performance.  At $N=100$, full scheduled
monitoring inspects 10 handoffs in the 10\% cartel regime and 20 in the 20\%
regime, normalized loads of only 0.101\% and 0.202\%, respectively.  Performance
under subsampled monitoring budgets is not evaluated here.

\begin{table}[!th]
\centering
\small
\begin{tabular}{rrrrrrrrr}
\toprule
Bidders & Cartel & Norm. load & HO groups & Seq. & Cont. & L1 & L2 & L3 \\
\midrule
25 & 10\% & 0.333\% & 2 & 0.917 & 0.500 & 0.500 & 1.000 & 0.500 \\
25 & 20\% & 0.667\% & 4 & 0.958 & 0.750 & 0.500 & 1.000 & 0.500 \\
50 & 10\% & 0.163\% & 4 & 0.958 & 0.750 & 0.500 & 1.000 & 0.500 \\
50 & 20\% & 0.408\% & 10 & 1.000 & 1.000 & 0.667 & 1.000 & 0.667 \\
75 & 10\% & 0.144\% & 8 & 1.000 & 1.000 & 0.667 & 1.000 & 0.667 \\
75 & 20\% & 0.252\% & 12 & 1.000 & 1.000 & 0.667 & 1.000 & 0.667 \\
100 & 10\% & 0.101\% & 10 & 1.000 & 1.000 & 0.667 & 1.000 & 0.667 \\
100 & 20\% & 0.202\% & 12 & 1.000 & 1.000 & 0.667 & 1.000 & 0.667 \\
\bottomrule
\end{tabular}

\caption{Three-layer VLA monitor results on the auction scale
grid, using held-out handoff-group splits within each
bidder-count/cartel-density setting. Latent-collusion rows are labeled positive;
independent, neutral, and blocked rows are labeled negative. Norm. load is the
number of monitored events divided by $N(N-1)$, expressed as a percentage; all
events in the deterministic handoff schedule are monitored.}
\label{tab:three-layer-scale}
\end{table}

For detector evaluation, each scheduled interaction contributes one row under
each of four matched conditions, for $4H=8\lfloor C/2\rfloor$ sidecar rows.
Independent and blocked rows are scheduled pair decisions even though no latent
is transmitted.  For example, $N=25$ and $\rho=10\%$ give $C=3$, $H=2$
scheduled interactions, and eight evaluation rows.  One cartel agent is
unpaired on each item; it still bids but contributes no paired row, and the
unpaired identity rotates across items.
``HO groups'' abbreviates ``held-out groups'' and is the number of successfully evaluated
leave-one-handoff-group-out folds, capped at 12.  A group contains the rows that
share an item, sender, and receiver---one row for each of the four matched
conditions.  Each fold holds out all four rows together.  Neutral rows from the
remaining groups fit the Layer~1 Gaussian and Layer~3 SAE and calibrate all three
layer thresholds.  Layer~2 does not train a classifier; it computes influence
from matched controls, with the held-out group's control rows available when
its suspect row is scored.  Thus, this layer evaluates a matched
counterfactual-access setting rather than control-free prospective detection.
Attack labels are used only to calculate held-out AUROC, which is averaged across folds.  The sequential gate remains high across the
10\% and 20\% cartel regimes, and Layer~2 is the dominant signal.

\subsection{Steering}

The steering grid separates prompt-level and activation-level interventions.
Blackbox steering modifies selected colluder prompts with market-compliance
instructions.  The system-level prompt applies a global behavioral instruction.
Whitebox steering applies an agent-level matched neutral counterfactual: selected
senders or unpaired cartel agents are rerun under neutral prompts, recomputing
their candidate-bid scores and outgoing latents; selected receivers instead
consume the sender's matched neutral latent before scoring bids.
Allocated steering applies whitebox edits to a fraction of selected cartel
agents and blackbox prompts to the remainder.  In
Table~\ref{tab:steering-outcomes}, ``WB/BB frac.'' reports these whitebox and blackbox fractions, respectively.  They sum to one for allocated,
blackbox, and whitebox policies; the system-only row is $0/0$ because its prompt is
global rather than assigned per agent.

We report two causal intervention effects relative to the matched unsteered
latent-collusion baseline.  ``$\Delta$ low bid'' is the percentage-point (pp) change
in the fraction of cartel bids at most 5 credits above the item's base price.  ``Dist.
recovery'' measures how far receiver bid distributions move from collusion
toward neutral.  For the set $\mathcal{K}$ of matched pair--item keys, define
\[
\bar D_x
:=
\frac{1}{|\mathcal{K}|}\sum_{u\in\mathcal{K}}
D_{\mathrm{JS}}\!\left(p_x^{(u)},p_{\mathrm{neutral}}^{(u)}\right),
\qquad
R_{\mathrm{dist}}:=1-\frac{\bar D_{\mathrm{int}}}{\bar D_{\mathrm{coll}}}.
\]
Here, $x\in\{\mathrm{int},\mathrm{coll}\}$ denotes the intervention or unsteered
latent-collusion condition, $p_x^{(u)}$ is the receiver bid distribution for key
$u$, and $D_{\mathrm{JS}}$ is the Jensen--Shannon divergence.  The table first
computes this ratio within each grid setting and then averages across settings.
Higher recovery is better, and 100\% denotes zero mean divergence from the
matched neutral distributions.

\begin{table}[t]
\centering
\small
\begin{tabular}{rrrrr}
\toprule
Campaign & Prompt & WB/BB frac. & $\Delta$ low bid & Dist. recovery \\
\midrule
steering\_allocated & blackbox & 0.25/0.75 & -16.2 pp & 34.7\% \\
steering\_allocated & blackbox & 0.5/0.5 & -28.3 pp & 70.1\% \\
steering\_allocated & blackbox & 0.75/0.25 & -39.0 pp & 93.0\% \\
steering\_blackbox & blackbox & 0/1 & -2.0 pp & 12.4\% \\
steering\_system & system & 0/0 & -29.5 pp & 2.9\% \\
steering\_whitebox & none & 1/0 & -47.3 pp & 100.0\% \\
\bottomrule
\end{tabular}

\caption{Steering outcomes and matched causal effects. Each row averages eight
grid settings: four bidder counts crossed with two cartel densities. Each
setting uses one seed and two auction items.}
\label{tab:steering-outcomes}
\end{table}

As the allocated whitebox fraction rises from 0.25 to 0.50 to 0.75,
receiver distribution recovery rises from 34.7\% to 70.1\% to 93.0\%, and the
cartel low-bid rate falls by 16.2, 28.3, and 39.0 percentage points.  Full
whitebox steering reaches 100\% distribution recovery and reduces low bidding by
47.3 percentage points.  Exact recovery is expected here because full whitebox
steering substitutes the same matched neutral reruns that define the recovery
target; it is therefore a counterfactual sanity check rather than evidence of
general attack removal.  The policies also affect the two metrics differently:
blackbox-only steering yields 12.4\% recovery and a 2.0-point low-bid reduction,
whereas the global system prompt yields a 29.5-point low-bid reduction but only
2.9\% distribution recovery.

\section{Discussion}

The controlled auction evidence supports three narrower claims about the
framework.  First, the homogeneous and heterogeneous experiments show that the
monitor discriminates the evaluated text- and fixed-latent-collusion conditions
across model families when matched controls are available.  Second, the
scalability grid shows that activation-aware monitoring can be evaluated on
scheduled pair decisions rather than requiring all-pairs activation capture.
Third, the steering grid characterizes how blackbox and whitebox interventions
change behavior relative to matched neutral controls.


The allocated steering design is practically relevant because real platforms may
have mixed permissions.  Hosted agents may permit whitebox activation access,
while third-party agents may only permit prompt-level blackbox intervention.  The
allocated grid tests this boundary directly: more whitebox coverage produces
monotonic improvements in bid-distribution recovery and low-bid reduction in
this grid.  Because whitebox steering directly replays matched neutral
counterfactuals, this dose response measures the benefit of increasing exact
counterfactual access under the benchmark assumptions.

\section{Related Work}

\paragraph{Latent multi-agent communication.}
LatentMAS demonstrates that agents can exchange continuous internal
representations instead of generating natural-language messages, substantially
reducing token use and inference time~\cite{zou2026latentmas}.  \shortmethod{}
addresses the complementary governance problem: how a host can audit and
intervene on a channel whose contents are absent from the public transcript.

\paragraph{Collusion and activation oversight.}
Prior work distinguishes tacit coordination, explicit cartels, and concealed
communication as threats in LLM-powered multi-agent systems
\cite{ghaemi2025collusion}.  \shortmethod{} combines three familiar forms of
evidence---representation anomaly, counterfactual behavioral influence, and
sparse-feature support---within an exact-join protocol that attributes public
actions to particular scheduled private-channel decisions.




\section{Conclusion}

Latent communication offers an efficient, scalable mode of coordination for
large multi-agent systems by reducing the need to serialize internal state into
long natural-language exchanges.  This benefit also
creates a governance gap: public transcripts reveal what agents do but may omit
the continuous states through which they coordinate.  \method{} offers a general
architecture for narrowing this gap: link private-channel records to public
actions, compare suspect decisions with matched controls, model neutral
communication rather than a fixed catalog of attacks, and choose interventions
according to the defender's level of model access.  The auction case study
provides a concrete proof of concept that this architecture can connect
detection, causal diagnosis, and scalable intervention in a multi-agent setting.


\end{document}